\documentclass[twoside]{article}

\usepackage{amsmath,amssymb,amsfonts,mathtools}
\usepackage{graphicx}
\usepackage{subcaption}
\usepackage{booktabs}
\usepackage{array}
\usepackage{marvosym} % Provides the \Letter symbol
\usepackage{stfloats}
\usepackage{color}
\usepackage{xcolor}

\usepackage[backend=biber,style=authoryear]{biblatex}
\usepackage{float}
\usepackage{algorithm}
\usepackage{algpseudocode}

\usepackage[preprint]{aistats2027}
\begin{document}



\twocolumn[

\aistatstitle{Bayesian Optimisation under State-Preservation Constraints}

\runningauthor{Diaz-Aylwin, Gopakumar, Myatra, Moulton, Zanisi, Leslie, Moss}

  \aistatsauthor{
    Gabriel Diaz-Aylwin  \And Vignesh Gopakumar
  }
  \aistatsaddress{
      Lancaster University, UK\\ \texttt{g.diaz-aylwin@lancaster.ac.uk} \And UK Atomic Energy Authority\\ \texttt{vignesh.gopakumar@ukaea.uk}
  }
  \aistatsauthor{
      Omkar Myatra \And David Moulton \And Lorenzo Zanisi
  }
  \aistatsaddress{
       UK Atomic Energy Authority \\ \texttt{omkar.myatra@ukaea.uk}\And UK Atomic Energy Authority \\ \texttt{david.moulton@ukaea.uk}\And UK Atomic Energy Authority\\ \texttt{lorenzo.zanisi@ukaea.uk}
  }
  \aistatsauthor{
       David S. Leslie\And Henry B. Moss 
  }
  \aistatsaddress{
       Lancaster University, UK \\ \texttt{d.leslie@lancaster.ac.uk}\And Lancaster University, UK \\ \texttt{henry.moss@lancaster.ac.uk}
  }
]

\begin{abstract}
In many engineering design problems, the objective and constraints depend on the state: the solution of a PDE determined by the design parameters. 
We consider improving a design while holding selected state observables near trusted values, which we call state preservation constraints. 
Constrained Bayesian optimisation handles these with a learnt feasibility model, but struggles with this problem's highly anisotropic feasible set. 
Our central idea is to pre-compute the set of controls whose linearised constraint response stays within tolerance, thereby pulling back the state-space constraint into design space. 
This linearisation defines an ellipsoid from which we can efficiently draw a large number of well-spread candidates. 
The underlying linear response map is refined online, and the ellipsoid is rebuilt accordingly. 
We demonstrate the method end-to-end on our key application - tokamak divertor optimisation under plasma-boundary preservation.
\end{abstract}

\section{Introduction}

% Figure 1
\begin{figure*}[t]\centering
\begin{subfigure}[b]{0.2699\textwidth}\includegraphics[width=\linewidth]{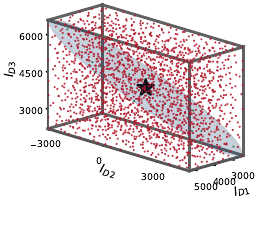}\caption{}\end{subfigure}\hfill
\begin{subfigure}[b]{0.2529\textwidth}\includegraphics[width=\linewidth]{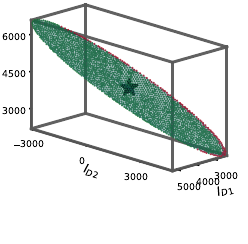}\caption{}\end{subfigure}\hfill
\begin{subfigure}[b]{0.1437\textwidth}\includegraphics[width=\linewidth]{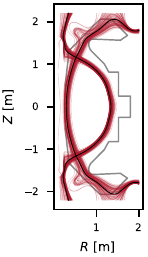}\caption{}\end{subfigure}\hfill
\begin{subfigure}[b]{0.1012\textwidth}\includegraphics[width=\linewidth]{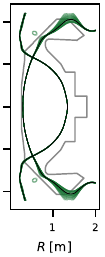}\caption{}\end{subfigure}\hfill
\begin{subfigure}[b]{0.1012\textwidth}\includegraphics[width=\linewidth]{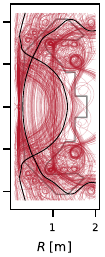}\caption{}\end{subfigure}\hfill
\begin{subfigure}[b]{0.1012\textwidth}\includegraphics[width=\linewidth]{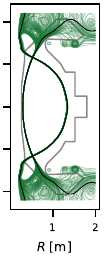}\caption{}\end{subfigure}
\caption{
Left: (a) uniform samples across coil-current design variables  $\mathcal X$ and (b) well-spread samples from the linearised feasible region $E$, restricted to the D1, D2, D3 coils for presentation.
Points feasible under the LCFS preservation constraint (desired fixed central region) are green, infeasible red. The starred point is our baseline solution.
Right: (c, d) the resulting PDE solutions of the sampled design variables superposed on the baseline solution (black) corresponding to the designs in (a) and (b). This is a vertical slice of a tokamak reactor torus.
(e, f) the same construction over the full $d=10$ current domain, showing how tightly $E$ constrains the LCFS.
Adopting $E$ as the proposal geometry within BO is the essence of our method.
}\label{fig:region}\end{figure*}

% ============================================================================================================
% Why is this an important problem?

% 1. Motivating Bayes Opt
There is a rich history of applying optimisation techniques to engineering design \parencite{forrester2008}.
Often, we rely on legacy, non-differentiable codebases to evaluate expensive objective functions \parencite{lam2018aerospace,du2023crash}.
% Here, \emph{sample efficiency} becomes the primary consideration.
\emph{Bayesian optimisation} (BO) is the natural tool for this regime, 
using a probabilistic surrogate to allocate a limited evaluation budget towards promising designs.
This framework has been applied to aerospace, structural and accelerator design \parencite{lam2018aerospace,duris2020fel,rostum2025bridge}.

% 2. Motivating state preservation:
In practical engineering design, however,
designs typically evolve from an established baseline.
Any improvements must then preserve the baseline's desirable properties:
an aerofoil re-profiled for lower drag must produce the lift the
aircraft was sized for \parencite{jameson1998navierstokes}
% a hull refined for lower resistance should displace the water the ship was designed to carry \parencite{campana2006hull};
and a blade reshaped for efficiency must admit the mass flow the engine was matched to \parencite{pierret1999blade}.
Here, checking a candidate design requires a full solve, ruling out simple rejection sampling.
The standard remedy in Bayesian optimisation is to learn
a surrogate of the constraint function from evaluations, and use this to guide subsequent experimentation \parencite{schonlau1998,gardner2014,gelbart2014}.
However, preservation constraints are a poor fit for this approach.
The preserved quantities are typically highly, and unevenly, sensitive to
the design parameters, making the feasible set a thin ellipsoid occupying
a vanishing fraction of the design space (Fig. \ref{fig:region}a).
Most proposed candidates are therefore infeasible,
and the surrogate must resolve this geometry from a handful of expensive scalar observations.

% The model must resolve a thin, anisotropic feasible region from a limited number of scalar observations.

% Our solution:
In this work, we instead introduce a classical construction from nonlinear
programming into constrained BO: we precompute the first-order approximation of
the feasible set and adopt it as a \textbf{constraint-aware proposal geometry} for the Bayesian optimisation loop.
This ellipsoidal region is available in
closed form, and millions of well-spread candidates can be drawn from
it at negligible cost (Fig. \ref{fig:region}b). We additionally refine its defining linear response map online by regression and rebuild the region accordingly.
Hence, at the cost of extracting a single Jacobian, we pre-compute and refine the geometry a purely data-driven method must spend its budget learning.

% Part 4: Application
Our primary application is \textbf{divertor design for tokamak fusion reactors} --- a critical problem plagued by expensive numerical physics simulations.
We seek to optimise a divertor's magnetic geometry for improved heat-exhaust performance while preserving the plasma boundary of a baseline equilibrium, known as the \textbf{last closed flux surface} (LCFS). This ensures the reactor's confinement properties are preserved (Fig. \ref{fig:region}c-f and Fig. \ref{fig:equilibria}).
Recent work applies Bayesian optimisation in this setting, using connection length as a performance proxy and learning core preservation as a constraint \parencite{bayes_opt_ukaea}.
We instead incorporate preservation directly into the proposal geometry, demonstrating substantial gains in sample efficiency over standard constrained-BO baselines, with ablations showing that the proposal geometry is the principal source of the improvement.

\section{Background}
% What is the reader going to need to understand in order to comperehend?

% Very broad background:
\paragraph{Bayesian optimisation.}
We wish to maximise an objective function $f:\mathcal X\to\mathbb R$.
We assume that $f$ is expensive to evaluate and derivative-free, making
sample efficiency our primary concern.
Bayesian optimisation addresses this setting by sequentially constructing a
probabilistic surrogate over the $n$ already observed evaluations
$\mathcal D_n$. %=\{(x_i,f(x_i))\}_{i=1}^n$. 
Gaussian processes (GPs) are widely used as surrogate models, providing a closed-form
posterior predictive distribution for $f(x)\mid\mathcal D_n$.
% \[
% f(x)\mid\mathcal D_n
% \sim
% \mathcal N\!\left(\mu_n(x),\sigma_n^2(x)\right)
% \]
\parencite{rasmussen2006}.

Each surrogate model is then used to choose the next evaluation, as selected by maximising an acquisition function
defined on this posterior. Acquisition functions balance regions of high predicted objective
value against those of high uncertainty, e.g. we can use the expected improvement (EI) \parencite{jones1998} over the score of the current best location (the incumbent)
\(
f_n^\star := \max_{i\le n} f(x_i),
\)
given by;
\[
\operatorname{EI}_n(x)
:=
\mathbb E_{f(x)|\,
\mathcal D_n}\!\left[
\max(f(x)-f_n^\star, 0)
\
\right].
\]

\paragraph{Constrained Bayesian optimisation.}
Suppose feasibility is defined by constraints $c_j(x)\leq 0$.
Standard constrained BO places probabilistic surrogates on the constraints and incorporates feasibility through the acquisition function;
for example, constrained expected improvement takes the form
\[
\operatorname{cEI}_n(x)
=
\operatorname{EI}_n(x)
\prod_{j} P_{c_j(x)\mid \mathcal D_n}\!\left[c_j(x)\leq 0\right],
\]
with the feasibility probability obtained from the learnt constraint models
\parencite{schonlau1998,gardner2014,gelbart2014}.
Alternatives use information-theoretic acquisitions or penalty-based formulations
\parencite{hernandezlobato2015pesc,lu2022epbo}. In all cases, feasibility is judged by a learnt constraint surrogate \parencite{eriksson2021scbo},
while the initial design and candidate generation remain over the ambient domain.
We show that for state preservation constraints these surrogates are often poor:
the feasible set is thin and anisotropic,
and a few expensive evaluations are not enough to resolve its boundary.

% For our optimisation to be applicable to industrial problems
% we must be capable of handling constraints.
% Suppose feasibility is defined by constraints $c_j(x)\leq 0$.
% Standard constrained BO places additional probabilistic surrogates on the constraints and incorporates feasibility through the acquisition function.
% For example, constrained expected improvement takes the form
% \[
% \operatorname{cEI}_n(x)
% =
% \operatorname{EI}_n(x)\,
% \mathbb \prod_{j} P_{c_j(x)\mid \mathcal D_n}\!\left[c_j(x)\leq 0
% \right],
% \]
% with the feasibility probability obtained from the learned constraint models
% \parencite{schonlau1998,gardner2014,gelbart2014}.
% Alternative approaches use information-theoretic acquisitions or penalty-based formulations
% \parencite{hernandezlobato2015pesc,lu2022epbo}.

% In standard constrained BO, the constraint surrogate enters primarily as a
% scoring mechanism to learn the probability of feasibility, with both the initial design points and subsequent candidate generation still performed over the ambient design domain. However, when feasibility occupies a thin, anisotropic region, such an ambient space-filling initial design may contain very few feasible points.
% The quality of the search is therefore limited throughout the optimisation by how accurately this surrogate 
% can resolve the complex feasible-set boundary from a small number of expensive evaluations. 
% Since the resulting constrained acquisition landscape may be highly non-convex, the geometry of the candidate set and the initialisation of local acquisition optimisation become critical.

\paragraph{Multi-objective Bayesian optimisation.}
% We make use of two standard extensions of Bayesian optimisation.
% First, trust-region BO localises search and modelling to an adaptively sized neighbourhood of the incumbent, improving robustness in higher-dimensional or heterogeneous objectives \parencite{eriksson2019turbo}.
In multi-objective BO (MOBO) it is common to replace seeking scalar improvements in the current best solution by instead seeking improvements over the problem's Pareto front --- the set of optimal tradeoffs between the multiple objectives. The resulting acquisition function is the Expected Hyper-Volume Improvement (EHVI) of \cite{emmerich2006}, implemented efficiently by \cite{daulton2020}.

% hypervolume improvement. Writing $\mathcal P_n$ for the current Pareto front and $r$ for a fixed reference point,
% \[
% \operatorname{HVI}_n(x)
% :=
% \mathrm{HV}\!\left(\mathcal P_n \cup \{f(x)\};r\right)
% -
% \mathrm{HV}\!\left(\mathcal P_n;r\right),
% \]
% and we use its posterior expectation,
% \(
% \operatorname{EHVI}_n(x)
% =
% \mathbb E\!\left[
% \operatorname{HVI}_n(x)
% \mid \mathcal D_n
% \right],
% \)
% to value progress on the Pareto front .

\paragraph{Local feasible-set approximations.} 
In this work, rather than learning feasibility pointwise, we will approximate the feasible set directly as a geometric object. 
This is a classical technique in constrained optimisation, however, this has yet to be fully explored in BO.
For example, sequential quadratic programming combines a local quadratic model of the objective with linearised constraints;
trust-region variants limit steps to a neighbourhood where these approximations remain reliable \parencite{nocedal2006} and 
derivative-free methods such as COBYLA construct local linear models of the constraints from function evaluations \parencite{powell1994}. 
We adapt these ideas from the nonlinear-programming literature 
and use the local linearised feasible region to provide candidate proposals within a Bayesian optimisation loop.
We further update this approximation of the feasible region online, using data observations.

\paragraph{Optimising under the simulator constraint.} 
One could avoid learning the constraint altogether and maximise the
acquisition function subject to the simulator-defined feasibility directly, making the inner loop a
PDE-constrained optimisation \parencite{hinze2008}. 
We do not --- this approach only makes sense when checking feasibility is significantly cheaper than evaluating the objective.
Furthermore, gradients may in principle be obtained by finite differences or adjoint methods, but, here, neither is straightforward.
Finally, gradient-based optimisation of the resulting non-convex acquisition function is typically local, leaving performance sensitive to initialisation --- 
the difficulty our proposal geometry addresses.

\section{Constraint-aware Candidate Generation}\label{sec:method}

Let $x\in\mathcal X\subset\mathbb R^d$ denote a design and let
$\psi(x)$ denote the corresponding system state produced by an
expensive simulator.
We seek to optimise an objective $F(\psi(x))$ while preserving a
selected observable $C(\psi(x))$ relative to a baseline design $x_0$.

We define feasibility by
\[
\|C(\psi(x))-C(\psi(x_0))\|_G\leq\tau,
\]
where $\|\cdot\|_G$ is a norm on the observable space and $\tau>0$
is a prescribed tolerance.
Writing
\[
v(x):=\|C(\psi(x))-C(\psi(x_0))\|_G,
\]
for the constraint violation.
Our optimisation problem is therefore
\[
x^\star\in
\operatorname*{arg\,max}_{x\in\mathcal X}
F(\psi(x))
\quad
\text{subject to}
\quad
v(x)\leq\tau.
\]

\subsection{The linearised feasible region}

The essence of our method is to approximate the feasible set to first order
and use this approximation as the proposal geometry for Bayesian optimisation.

Around the baseline design $x_0$, a first-order expansion gives
\[
C(\psi(x_0+\delta x))
=
C(\psi(x_0)) + J\,\delta x + O(\|\delta x\|^2).
\]
Where the Jacobian 
\(
J := D(C\circ\psi)(x_0).
\)
Hence
\[
C(\psi(x_0+\delta x)) - C(\psi(x_0))
\approx
J\,\delta x,
\]
and therefore
\[
v(x_0+\delta x)^2
\approx
\|J\,\delta x\|_G^2
=
\delta x^\top J^\top GJ\,\delta x .
\]
Writing
\(
M := J^\top GJ,
\)
the first-order approximation to the preservation constraint
$v(x_0+\delta x)\leq\tau$ becomes
the \emph{linearised feasible region}
\[
E :=
\left\{
x_0+\delta x:
\delta x^\top M\,\delta x \leq \tau^2
\right\}.
\]
We use $E\cap\mathcal X$ as the proposal geometry for the Bayesian optimisation loop.
The intersection keeps the region bounded when $M$ is singular.

This construction admits a geometric interpretation.
The inner product $G$ measures changes in the observable space, while
$J$ pulls this structure back to design space through
\[
M = J^\top GJ.
\]
Thus $E$ consists of design perturbations whose predicted first-order
observable displacement is bounded by $\tau$.

\subsection{Ray-based intersection sampling and maximin filtering}

At each stage of the optimisation we draw candidates from
\(
E \cap B,
\)
where $E$ is the linearised feasible region and $B$ is either the
engineering domain $\mathcal X$ or the current objective-focused trust region. Because $E$ may be highly anisotropic, naive box sampling can give poor coverage.
We therefore generate a large candidate pool by ray-based sampling within
$E\cap B$, using analytic intersections with the ellipsoidal and box
boundaries, and then greedily retain a well-spread subset using a maximin
criterion \parencite{johnson1990minimax}.
This procedure allows very large candidate pools to be generated cheaply;
implementation details are given in Appendix~\ref{app:sampling}.

\subsection{Online refinement}

The initial response map
\[
J = D(C\circ\psi)(x_0)
\]
describes the infinitesimal change in the observable around the baseline.
As the optimisation explores finite displacements from $x_0$, nonlinear effects
cause this local differential to become an imperfect predictor of the observed
change $C(\psi(x))-C(\psi(x_0))$.

Every simulator evaluation, however, provides additional information about this
response at no extra solver cost. Writing
\[
\delta x_i := x_i-x_0,
\qquad
\delta C_i := C(\psi(x_i))-C(\psi(x_0)),
\]
each evaluated design supplies a displacement pair
$(\delta x_i,\delta C_i)$ satisfying approximately
\[
\delta C_i \approx \widehat J\,\delta x_i.
\]
We therefore refit the linear response map $\widehat J$ by least squares over
the pairs observed so far, including the initial finite-difference
probes.
The proposal geometry is then updated through
\[
\widehat M=\widehat J^\top G\widehat J,
\qquad
E=
\left\{
x_0+\delta x:
\delta x^\top\widehat M\delta x\leq\tau^2
\right\}.
\]
Thus the finite-difference Jacobian supplies an initial proposal geometry before
BO has collected data, while subsequent evaluations continually
adapt that geometry to the finite-scale behaviour encountered by the search.

\subsection{Feasibility beyond the linearisation}\label{sec:feasibility}

Due to linearisation error, the edges of our feasible region will contain
some infeasible points. To handle these, we use a learnt feasibility model to weight the
acquisition. Rather than model the scalar violation, we model the change in the
observable, $\delta C(x)$ and take its norm:
\[
\delta C(x) = \hat J\,\delta x + \sum_{j=1}^{K} a_j(x)\,\phi_j,
\]
where $\hat J$ is the online-refit linear response map from which $E$ is built,
$\{\phi_j\}$ are the leading $K$ modes of an uncentred proper orthogonal
decomposition of the residual fields $\delta C_i - \hat J\,\delta x_i$ and the $a_j$ are
zero-mean GPs on the mode coefficients, learning what is missed by the linearisation \parencite{kennedy2001}. 
Since $\delta C(x)$ is Gaussian under this model, $\lVert\delta C(x)\rVert_G^2$
is a quadratic form of a Gaussian vector and $\mathbb P[\, \lVert\delta C(x)\rVert_G \le\tau \,]$ is a
generalised $\chi^2$ tail, which we approximate by moment matching.

We refer to this model as our \textbf{field model}, since $C$ is
field-valued. The construction --- a low-rank basis for the field with
GPs on its coefficients --- is standard for functional outputs
\parencite{higdon2008,guo2018} and, as we show in Section \ref{sec:experiments}, outperforms fitting a scalar GP directly on $\lVert\delta C(x)\rVert_G$.

% our method only needs an estimate of 
% $\mathbb P[\, \lVert\delta C(x)\rVert_G \le\tau \,]$ to capture any residual nonlinearity and so  we ablate it with a  ().

\section{Divertor Optimisation}

\begin{figure}[t]\centering
\begin{subfigure}[b]{0.3721\columnwidth}\includegraphics[width=\linewidth]{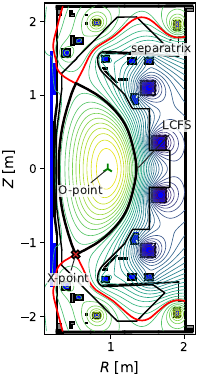}\caption{}\end{subfigure}\hfill
\begin{subfigure}[b]{0.3020\columnwidth}\includegraphics[width=\linewidth]{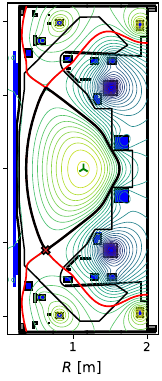}\caption{}\end{subfigure}\hfill
\begin{subfigure}[b]{0.3020\columnwidth}\includegraphics[width=\linewidth]{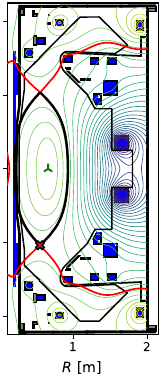}\caption{}\end{subfigure}
\caption{(a) A reconstructed super-X equilibrium from MAST-U shot $53349$. We take this equilibrium as our baseline (starting design).
(b, c) Random current perturbations (within known MAST-U PF coil bounds) demonstrating the sensitivity of the LCFS (black traced contour). The centre of the black boxes denotes coil location. We will be optimising the current strengths (and later the coil locations) in order to get desirable properties of the separatrix (red contour line), whilst attempting to preserve the LCFS at its baseline shape.
}\label{fig:equilibria}\end{figure}

We now explain our key application of divertor optimisation, as summarised in Fig. \ref{fig:equilibria}.

% FIGURE 4 -- baseline equilibrium + four box draws, one PDF per panel.
\label{fig:eqs}

% Explain fusion (up to separatrix)

\paragraph{Design $x$.}
Tokamak fusion reactors confine a high-temperature plasma
with a magnetic cage of helical field lines.
This magnetic field is induced by current flowing through external coils
and through the plasma itself.
The \textbf{poloidal field coils} shape the confined plasma and the divertor geometry.
We take as design $x = I$, the poloidal field coil currents, and later $x = (I, R, Z)$ with their positions.

\paragraph{State $\psi$.}
Under axisymmetry, the divergence-free magnetic field admits a scalar
representation in terms of a single function:
\(
\psi(R,Z)
\)
-- the \textbf{poloidal magnetic flux function} and constitutes our state.
The level sets of \(\psi\) define magnetic flux surfaces.
At ideal MHD equilibrium, both the pressure and toroidal field function are constant on flux surfaces
and the force-balance equation
reduces to the Grad-Shafranov equation
\[
\Delta^\star \psi
=
-\mu_0 R^2 p'(\psi)
-
F(\psi)F'(\psi),
\]
where
\[
\Delta^\star
=
R\frac{\partial}{\partial R}
\left(
\frac1R
\frac{\partial}{\partial R}
\right)
+
\frac{\partial^2}{\partial Z^2}
\]
is the Grad-Shafranov operator.
% Maybe need to motivate free boundary
In this work, we use
\emph{FreeGSNKE} \parencite{amorisco2024}, to solve the free-boundary Grad--Shafranov equation (GSE).
The GSE takes as inputs the currents and positions of the poloidal field coils
and returns the poloidal flux function over the full spatial domain. Example such equilibria are plotted in Figure 2.

\paragraph{Constraint $C$.}
Charged particles follow magnetic field lines.
The field lines arising from solving the GSE may be open, so the charged particles can interact with material surfaces, or closed, when they are confined.
The boundary between closed and open field lines is known as the separatrix.
% the level set of a hyperbolic critical point of $\psi(R, Z)$, the \emph{X-point}.
On this flux surface, the point(s) where the poloidal magnetic field goes to zero is called the X-point. The \textbf{last closed flux surface} (LCFS) (Figure 2) is the outermost closed magnetic flux surface.
The shape of this closed curve is chosen to optimise plasma confinement and stability requirements and therefore acts as our observable $C$.

For a smooth curve, first-order shape changes are described by normal displacement \parencite{Michor2006,sokolowski1992}.
Let $\Gamma_0$ denote the baseline LCFS. We represent the change to the perturbed LCFS 
$\Gamma(x)$ by $\delta C(x)\in\mathbb{R}^N$, whose entries are signed normal displacements measured at $N$ equally spaced arclength locations on $\Gamma_0$ (construction in Appendix~\ref{app:fd}).

Finite differences with respect to the design variables give the response map
\[
J=D(C\circ\psi)(x_0)\in\mathbb{R}^{N\times d},
\]
a matrix whose $j$th column is the LCFS normal-displacement field produced by a small
perturbation of design variable $x_j$. 
We measure preservation by the RMS normal displacement,
\[
\begin{aligned}
v(x) &:= \|\delta C(x)\|_G
       = \left(\frac{1}{N}\,\delta C(x)^\top\delta C(x)\right)^{1/2}, \\
\|y\|_G &:= \bigl(y^\top G\,y\bigr)^{1/2}, \quad\text{i.e. } G = \tfrac{1}{N} I.
\end{aligned}
\]

\paragraph{Objective $F$.}
A fusion plasma is not a closed system -- charged particles and heat
continuously escape the confined core and intersect material
surfaces on the tokamak wall.
In reactor-scale tokamaks, such unmitigated exhaust heat loads far exceed the engineering limits of any practical plasma-facing material \parencite{kotschenreuther2007heat}.
To manage this, tokamaks employ a dedicated region known as the \emph{divertor}.
The central objective of the divertor is to achieve and control a \emph{detached} plasma
regime, characterised by sufficiently strong power and momentum losses that the plasma cools and begins
to recombine before reaching the divertor target. The result is a large reduction of the parallel heat flux incident on material surfaces.

We minimise the \textbf{detachment-threshold impurity fraction, $c_z$}: the impurity fraction required for detachment onset at the divertor target, at fixed upstream conditions. 
Lowering this threshold makes detachment easier to access and widens the operating window \parencite{Lipschultz2016}. 
We compute this objective using the reduced divertor model \emph{FusionDLS} \parencite{Cowley2022, dls_extended}.

\subsection{Related work in Fusion Research}

%\paragraph{Linearisations for plasma shape control}
We note that considering the \textbf{linearised coil-to-boundary response} is a well-studied construct in tokamak magnetic control.
\emph{Virtual circuits} use it to vary one geometric shape parameter while holding the others fixed \parencite{MCARDLE2020111764, Anand_2024}.
We use the same principle to construct a search region rather than control directions.
Relaxing the response's null space to a region would require a norm on shape parameter changes, instead, we use the $L^2$ norm of the LCFS normal displacement.

%\paragraph{Core preservation.}
Two additional constructions for \textbf{holding the core fixed} appear in the literature and we present them below. However, in both, note that constraint approximations are controlled coarsely by a discrete parameter that sacrifices approximation quality for degrees of freedom. In contrast, our norm-based constraint decouples LCFS resolution from design dimension:
we are able to resolve the LCFS at $\mathcal{O}(10^4)$ points while retaining all $d$ design directions, with $\tau$ controlling the allowed displacement.

\begin{enumerate}
    \item \textcite{Bardsley_2024} preserve the core approximately by matching the leading spherical-harmonic coefficients 
of $\psi_{\mathrm{machine}}$ to their baseline values. These coefficients are linear in coil currents, so admissible current changes lie in the null space of the resulting constraint matrix.
% Critique:
Each additional harmonic tightens
preservation at the cost of a degree of freedom,  with no intermediate
settings available. The approximation is also conservative as the constraint fixes the field throughout a source-free ball containing the plasma rather than directly preserving the LCFS. 
%For this approximation to be valid, any coil lying inside the ball must be held fixed.
\item 
\textcite{forge} place conditions on the machine flux at sampled points along the
separatrix to define a linear system whose null space supplies
simulated-annealing proposals that satisfy the
constraints. However, the exactness is only enforced pointwise, with the boundary pinned
at a set number of chosen points and unconstrained between them.
\end{enumerate}

\section{Experiments}
\label{sec:experiments}
% \subsection{Setup}

\paragraph{Experimental protocol.}
All experiments start from the same MAST-U-like baseline equilibrium with LCFS-displacement tolerance $\tau=0.5$ cm.
We first optimise $d=10$ paired coil currents, then jointly optimise currents and positions ($d=30$), and finally consider an additional objective function. 
Evaluation budgets are $1{,}000$ (currents-only) and $1{,}500$ (joint), with $50$ and $100$ initial designs respectively.
Constructing the initial response map requires an additional $2d$ finite-difference solves. %, excluded from these budgets.
We report $50$ seeds per run, each on $4$ CPU cores with $16$~GB and $40$~GB of memory for $d=10$ and $d=30$. See Appendix~\ref{app:hyper}.

\paragraph{Shared implementation details.}
% I think this is too strong. 
We compare our proposed algorithm with a variant of the previous application of BO to divertor optimisation \parencite{bayes_opt_ukaea} (henceforth, the competitor). These two methods differ principally (see Appendix~\ref{app:hyper}) in candidate generation and feasibility modelling ---  \cite{bayes_opt_ukaea} attempt to learn core preservation (with a Matérn-$5/2$ GP on the scalar violation $v(x)$) whereas we approximate the feasible set directly as a geometric object (details below). For both methods, objectives are modeled with zero-mean GPs with ARD Matérn-$5/2$ kernels, fit by maximum likelihood.
Candidates are ranked by feasibility-weighted expected improvement,
$\operatorname{EI}(x)\mathbb \times P[v(x)\le\tau]$ %\parencite{schonlau1998,gardner2014},
relative to the best observed feasible value.
The highest-ranked candidates initialise L-BFGS-B refinement of this acquisition. Both methods deploy the Trust-Region BO (TuRBO) of  \cite{eriksson2019turbo} to localise search and modelling to an adaptively sized neighbourhood $B$ of the incumbent. Upon trust-region convergence, the objective GP is reset and a fresh initial design is drawn.

%within a TuRBO box $B$ \parencite{eriksson2019turbo}.
%On restart, the objective GP is reset and a fresh initial design is drawn; engineering bounds $\mathcal X$ remain fixed.

\paragraph{Our algorithm.}\label{sec:ours}
The primary contribution of our paper is the \emph{continually updating candidate proposal geometry}:
the initial design and candidate pools are drawn from the linearised
feasible region $E$ of \S\ref{sec:method}.
At each iteration, we draw $10^5$ candidates from $E\cap B$ using ray-based sampling and apply maximin filtering.
Acquisition may leave $E$, with feasibility assessed by the field model of \S\ref{sec:feasibility}, using $K=5$ modes.
After each step, we refit $\hat J$ on all observed pairs, rebuild $E$, and update surrogates and trust regions, see Algorithm \ref{alg:bo}.
%The response map and feasibility model retain all observations across trust-region restarts.

% % Forced placement; reduced spacing and font size to fit the column.
% {\setlength{\intextsep}{4pt}
\begin{algorithm}[]\small
\caption{Constrained BO with linearised proposal geometry (\# initial designs $N_{init}$, \# steps $T$)}
\label{alg:bo}
\begin{algorithmic}[1]
\State Solve $\psi_0=\psi(x_0)$
\State Estimate $\hat J$ by finite differences ($2d$ solves)
\State $E \gets \{x_0+\delta x:\delta x^\top \hat J^\top G\hat J\,\delta x\le\tau^2\}$; \quad $B\gets\mathcal X$
\State Sample $N_{\rm init}$ points from $E\cap\mathcal X$; evaluate $F,\delta C$
\State Refit $\hat J$, $E$, and surrogate models
\For{$t=1,\ldots,T$}
    \State Sample candidates from $E\cap B$; 
    \State Maximin filter candidates; rank by $\mathrm{EI}\,\mathbb P[v\le\tau]$
    \State Refine top candidates with L-BFGS-B; select $x^*$
    \State Evaluate $F(\psi(x^*))$, $\delta C(x^*)$
    \State Refit $\hat J$, $E$, and surrogate models
    \State Update, rescale, and recentre TuRBO region $B$
    \If{$|B|<L_{\min}$}
        \State Restart objective GP 
        \State Sample $N_{\rm init}$ points from $E\cap\mathcal X$; evaluate $F$
        \State Initialise new TurBO region $B$
    \EndIf
\EndFor
\State \Return best feasible design
\end{algorithmic}
\end{algorithm}
% }

\subsection{Experiment 1: currents only (d=10)}\label{sec:currents}

\begin{figure*}[t]\centering
\begin{subfigure}[b]{0.2469\textwidth}\includegraphics[width=\linewidth]{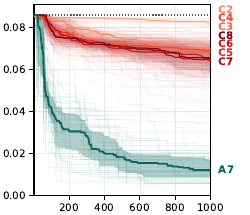}\caption{}\label{fig:res-a}\end{subfigure}\hfill
\begin{subfigure}[b]{0.2387\textwidth}\includegraphics[width=\linewidth]{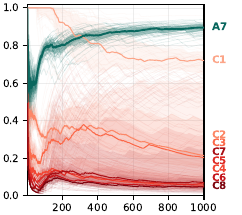}\caption{}\label{fig:res-b}\end{subfigure}\hfill
\begin{subfigure}[b]{0.2469\textwidth}\includegraphics[width=\linewidth]{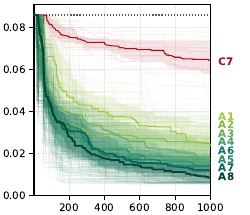}\caption{}\label{fig:res-c}\end{subfigure}\hfill
\begin{subfigure}[b]{0.2407\textwidth}\includegraphics[width=\linewidth]{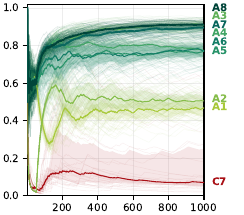}\caption{}\label{fig:res-d}\end{subfigure}
\caption{Currents-only ($d=10$), 1000 evaluations. Median (bold), IQR
(band) and every individual seed (faint). \textbf{(a),(c)} y-axis is best feasible
$c_Z$; the dotted line is the baseline $c_Z=0.086$. \textbf{(b),(d)} y-axis is
cumulative feasible fraction. \textbf{(a),(b)} our method (A7) against
the full competitor sweep C1--C8 over its trust-region size $L$.
\textbf{(c),(d)} the best competitor setting (C7) against every ablation
A1--A8. Identifiers and colours are shared across all four panels and
with Table~\ref{tab:results}.}
\label{fig:results}\end{figure*}

% Generated, never retyped:
%   .venv/bin/python -m figures.make_results_table > results_table.tex
% Per-row seed counts are written to results_table.log as provenance.
\begin{table*}[t]\scriptsize\setlength{\tabcolsep}{3.5pt}
\begin{minipage}[c]{0.655\textwidth}\centering
\newcommand{\B}[1]{\textbf{#1}}

\begin{tabular}{@{}llcccc@{}}
\toprule
& Run & median $c_Z$ & IQR & median feas. & IQR \\
\midrule
C1 & competitor, $L=0.001$ & 0.0827 & [0.0787, 0.0837] & 0.720 & [0.646, 0.768] \\
C2 & competitor, $L=0.005$ & 0.0702 & [0.0665, 0.0728] & 0.214 & [0.155, 0.619] \\
C3 & competitor, $L=0.01$  & 0.0683 & [0.0651, 0.0723] & 0.203 & [0.083, 0.398] \\
C4 & competitor, $L=0.015$ & 0.0692 & [0.0653, 0.0717] & 0.057 & [0.045, 0.103] \\
C5 & competitor, $L=0.02$  & 0.0646 & [0.0563, 0.0722] & 0.064 & [0.039, 0.321] \\
C6 & competitor, $L=0.025$ & 0.0655 & [0.0622, 0.0696] & 0.049 & [0.033, 0.091] \\
C7 & competitor, $L=0.03$  & 0.0636 & [0.0582, 0.0726] & 0.071 & [0.053, 0.251] \\
C8 & competitor, $L=0.035$ & 0.0657 & [0.0577, 0.0702] & 0.047 & [0.034, 0.093] \\
\midrule
A1 & scalar GP, unconfined & 0.0244 & [0.0209, 0.0334] & 0.468 & [0.429, 0.503] \\
A2 & box proposals         & 0.0238 & [0.0137, 0.0283] & 0.506 & [0.469, 0.525] \\
A3 & geometry only         & 0.0178 & [0.0120, 0.0244] & 0.908 & [0.877, 0.929] \\
A4 & frozen map            & 0.0155 & [0.0110, 0.0203] & 0.774 & [0.748, 0.815] \\
A5 & scalar GP, confined   & 0.0140 & [0.0099, 0.0198] & 0.769 & [0.734, 0.800] \\
A6 & hard gate             & 0.0142 & [0.0106, 0.0186] & 0.772 & [0.736, 0.800] \\
\B{A7} & \B{full}   & \B{0.0122} & [0.0090, 0.0171] & \B{0.895} & [0.876, 0.908] \\
\B{A8} & \B{no box} & \B{0.0087} & [0.0069, 0.0134] & \B{0.911} & [0.895, 0.920] \\
\bottomrule
\end{tabular}
\end{minipage}\hfill
\begin{minipage}[c]{0.33\textwidth}\centering
\fbox{\parbox{\dimexpr\linewidth-2\fboxsep-2\fboxrule}{%
\begin{tabular}{@{}l@{\ \ }>{\raggedright\arraybackslash}p{\dimexpr\linewidth-2.4em}@{}}
A1 & proposals from $E$; competitor's scalar GP; acquisition optimiser can leave $E$\\[1pt]
A2 & competitor's box proposals ($L = 0.03$, as C7); our field model (mean $\hat J\,\delta x$)\\[1pt]
A3 & proposals from $E\cap B$; no objective or constraint surrogate; random selection\\[1pt]
A4 & A7 with $\hat J$ frozen at the finite-difference Jacobian\\[1pt]
A5 & A1 with the optimiser held inside $E$\\[1pt]
A6 & A5 with the scalar GP removed; $E$ alone decides feasibility\\[1pt]
A7 & the full algorithm \ref{alg:bo}\\[1pt]
A8 & A7 with the box trust region removed
\end{tabular}}}
\end{minipage}
\caption{Final performance at 1000 evaluations. Identifiers match
Fig.~\ref{fig:results}; the box defines the ablation.}
\label{tab:results}\end{table*}

We first consider the task of optimising currents to minimise the detachment-threshold impurity fraction. Results are presented in Figure ~\ref{fig:results} and Table~\ref{tab:results}, alongside a set of ablations on our method (A1-A8) and the competitor (C1-C8) to 
separate the effects of proposal geometry, feasibility modelling, online refinement and the box trust region. These results show substantial gains over the existing method and raise the following observations. 

\paragraph{Our proposal geometry is the principal source of improvement.}
Most of the performance gain is already obtained once this geometry is used
for proposals. A3 uses \textbf{no surrogate of any kind} (sampling randomly from $E\cap B$), yet reaches a median final
$c_Z$ of $0.018$ at $90.8\%$ feasibility --- $3.6\times$ better than the
best competitor setting. It also outperforms both methods that adopt only
one of our other components: the field model with box proposals (A2) and the scalar GP with proposals from $E$ but unconfined
acquisition refinement (A1).

\paragraph{Online refinement is important for maintaining feasibility.}
Online refitting is primarily a feasibility correction as the local
linearisation degrades. Freezing the response map (A4) costs roughly
$12 \%$ feasibility with no significant objective effect.

\paragraph{The field model allows us to safely leave the linearised
region.}  While unconstrained refinement under the scalar GP gives $47\%$ feasibility (A1),
hard confinement to $E$ raises it to  $77\%$ (A5, A6). Adding the field model and letting L-BFGS refinement leave E improves this further to
$89.5\%$ (A7). These excursions, however, buy a minimal amount of objective improvement:
A5 to A7 moves the median $c_Z$ by a couple percentage points from the baseline, pointing again to the geometry as the key driver.

\paragraph{Trust regions are critical for the competitor, but not for us.} Removing trust regions (A8) does not hurt our algorithm, and here even improves it, but makes the competitor a non-starter --- sampling and acquisition optimisation over the full engineering bounds yield almost no feasible points and many equilibrium-solver failures. Moreover, even when using trust regions within the competitor, the initial side length of the trust region must be carefully chosen to be small enough to find feasible designs, but large enough to permit useful optimisation.
We sweep eight initial side lengths $L$ (C1--C8) and see that
% At $L=0.001$ the competitor stays $72\%$ feasible but improves on the
% baseline $c_Z$ by only $4\%$.
larger $L$ buys progress at the cost of feasibility -- once $L\ge0.015$, feasibility falls to below $8\%$ and progress stalls at $c_Z\approx0.065$.
Every competitor setting is markedly less sample efficient and has worse cumulative feasibility than our method (A7/A8) -- we reach a median final $c_Z$ of $0.012$, $5.2\times$ better than the best competitor setting, while keeping $89.5\%$ of its evaluations feasible.

% $c_Z$ from $0.012$ to $0.009$, with a slight increase in feasibility.

%For the joint problem ($d=30$, \S\ref{sec:joint}), objective performance is similar with and without the box, while feasibility increases from $67.9\%$ to $83.3\%$.
%We view the trust region in our full algorithm as a problem-specific choice, and keep it for the improved joint feasibility.

\subsection{Experiment 2: include positions (d=30)}\label{sec:joint}

% FIGURE 5 -- combined mode, five panels in ONE row.
\begin{figure*}[t]\centering
\begin{subfigure}[b]{0.2676\textwidth}\includegraphics[width=\linewidth]{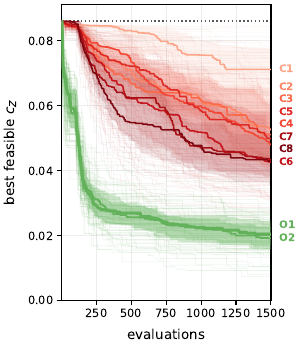}\caption{}\end{subfigure}\hfill
\begin{subfigure}[b]{0.2583\textwidth}\includegraphics[width=\linewidth]{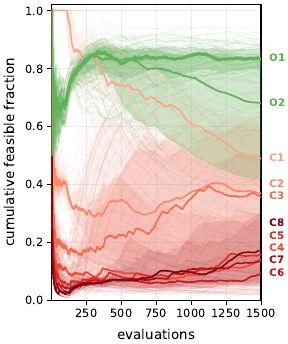}\caption{}\end{subfigure}\hfill
\begin{subfigure}[b]{0.1754\textwidth}\includegraphics[width=\linewidth]{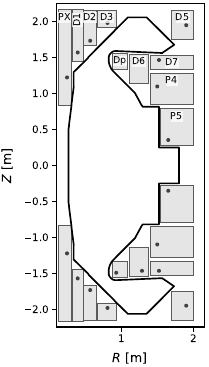}\caption{}\end{subfigure}\hfill
\begin{subfigure}[b]{0.1333\textwidth}\includegraphics[width=\linewidth]{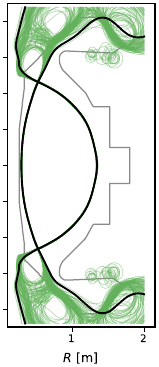}\caption{}\end{subfigure}\hfill
\begin{subfigure}[b]{0.1333\textwidth}\includegraphics[width=\linewidth]{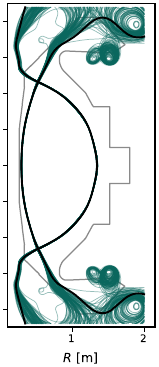}\caption{}\end{subfigure}
\caption{Left: Currents and position combined optimisation plots: (a) best $c_Z$-so-far and  (b) cumulative feasibility, vs our competitor method (C1-C8) .
O1: joint with box trust region, O2: without.
Right: (c) An illustration of the position bounds, (d) our optimal feasible separatrices from the combined optimisation and (e) from our currents-only optimisations. }\label{fig:combined}\end{figure*}

So far, the design variables have been the PF-coil currents alone.
We now extend the control vector to include coil positions,
\(
x=(I,R,Z)\in\mathbb R^{d_I+2d_R}.
\)
This combined optimisation stresses our method as its control to LCFS perturbation response is much more nonlinear than with currents-only. Each control is normalised by its engineering bounds, 
and the response map $J$ is computed in these normalised coordinates.
%The proposal-region construction is otherwise unchanged.
The bilinear structure of the equilibrium response to positions, and the coordinate transformation it motivates, are deferred to Appendix~\ref{app:bilinear}. Fig.~\ref{fig:combined} shows that our method is markedly more sample-efficient than the competitor, achieving a $2.1\times$ lower detachment threshold at the same evaluation budget. While our method is robust to trust region removal (O1-2), the competitor remains sensitive to trust region setup (C1-8).

% The added position controls perturb the boundary less per unit of normalised displacement, improving feasibility for box proposals.
% Our method instead worsens, suggesting that the increased search dimension outweighs the added design freedom at this budget.
% Nevertheless, it remains 

% Joint optimisation narrows the gap: our median final $c_Z$ is $0.020$, versus the best competitor's $0.043$ (Fig.~\ref{fig:combined}).

\paragraph{Online refinement is critical as we increase non-linearity }
Coil positions enter the equilibrium more nonlinearly than currents.
With positions included, the frozen linearisation's median violation-prediction error is $1.07$ cm, exceeding the $0.5$ cm tolerance.
Online refitting reduces the one-step-ahead (held-out) median prediction error to $0.031$ cm, allowing us to retain $83.3\%$ feasibility despite stronger nonlinearity.

\subsection{Extension 3: an additional objective}

% FIGURE (MOBO) -- one TEXT COLUMN, arms superposed.
\begin{figure}[t]\centering
\begin{subfigure}[b]{0.5018\columnwidth}\includegraphics[width=\linewidth]{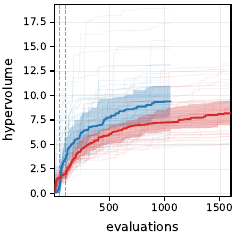}\caption{}\end{subfigure}\hfill
\begin{subfigure}[b]{0.4842\columnwidth}\includegraphics[width=\linewidth]{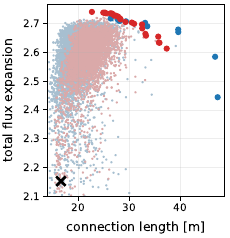}\caption{}\end{subfigure}
\caption{
    Left: HV plots from currents-only MOBO and currents and positions MOBO.
    Right: The corresponding connection length and total flux expansion pareto plots.
    Red are the joint MOBO results, blue the currents-only. The `x' is our baseline equilibrium and the HV anchor.
}\label{fig:mobo}\end{figure}

The focus of our paper is on the handling of preservation constraints,
and our exact same machinery can also be applied to multi-objective Bayesian optimisation by swapping EI with EHVI. Following \textcite{Cowley2022}, we seek to jointly optimise with respect to two key quantities that govern the
controllability of a detachment front: \emph{connection length} $L_c$
(field-line distance from midplane to target) and \emph{total flux
expansion} (ratio of field strength at the X-point to that at the target).
%Both widen the detachment window, so we wish to maximise them jointly. 
Fig.~\ref{fig:mobo} demonstrates effective MOBO, finding dense Pareto fronts for both the currents-only (blue) and currents and positions (red) MOBO. We do not run a competitor here, as this would require a non-trivial multi-objective trust-region. The point of this demonstration is to show that our proposal geometry is indifferent to the choice of acquisition function.

% \paragraph{Our constraint construction leads to improved Pareto fronts}
% Both methods move the design a long way from the baseline in both objectives (Fig.~\ref{fig:mobo}).
% % Against a baseline of $L_c = 16.6$ m and a total flux expansion of $2.15$, the currents-only front reaches $L_c$ up to $47.3$ m ($2.9\times$) and flux expansion up to $2.72$ ($1.3\times$), and the joint front up to $37.3$ m and $2.74$.
% Feasibility is in fact higher than in the single-objective runs ($96.8\%$ currents-only, $98.0\%$ joint)
% -- the multi-objective acquisition does not drive as hard into the constraint boundary.
% The currents-only arm attains the larger hypervolume and the longer connection lengths, while the joint arm returns a denser front and marginally higher flux expansion.
% The pattern mirrors the single-objective case: the extra twenty dimensions have not yet paid for themselves at this budget. Note that we have no competitor to run here, as this would require a non-trival multi-objective trust-region.  The point of this demonstration is to show that our proposal geometry is indifferent to the choice of acquisition function and so can be applied across a wide range of BO settings.

\section{Conclusion}

Preservation constraints admit a thin, anisotropic feasible set that is
expensive to learn but cheap to compute.
Here, uniform sampling of the $d = 10$ current box yielded no feasible designs in $1{,}000$ solves (Fig.~\ref{fig:region}e). Thus, a data-driven method must locate the set from scalar observations that almost never land inside it. We get around this by precomputing its
first-order approximation from $2d$ solves and refining it online. On the divertor optimisation problem, using this geometry to propose candidate designs inside a Bayesian optimisation loop
reaches a median detachment threshold $5$--$7\times$ below the best competitor in the currents-only problem while holding $\sim\!90\%$ of evaluations within a
$0.5$ cm boundary tolerance.%, and the ablations attribute the gain to the
% geometry rather than the surrogate: the same field model proposing from a
% box (A2) loses to no surrogate at all proposing from $E$ (A3).

We believe our approach provides a general framework for the Bayesian optimisation of trusted engineering designs under tightly constrained state variation. Future work will apply our method to other key problems in heavily-constrained and high-cost engineering, like radio-frequency cavity \parencite{wang2026multiobjective} and stellarator \parencite{packman2025bayesian} design. Moreover, as our experiments show on MOBO, the proposed BO module is self-contained and can be easily plugged into orthogonal BO advances, including batch \parencite{gonzalez2016batch}, multi-fidelity \parencite{snoek2012practical}, multi-task \parencite{moss2020mumbo} and high-dimensional \parencite{doumont2025we} BO, as well as BO under risk-adversity \parencite{picheny2022bayesian}.

% \clearpage
% \subsection*{AI Use Statement}
% Our use of generative AI in this paper was limited to the following: obtaining feedback on methodology and experiments, implementing methods, generating synthetic data, creating scientific figures, writing and editing code, brainstorming, and finding information from external sources. No generative AI tools were involved in constructing theoretical models or conceptual frameworks, forming or refining hypotheses, translation, dataset cleaning or reformatting, qualitative or thematic analysis of data, or the interpretation of results. Every AI-assisted output was reviewed by the authors, and all LLM-generated code was validated through testing. The authors remain fully responsible for the content of this work, including any text, claims, or artifacts developed with generative AI assistance.

\printbibliography

\appendix
\onecolumn
\aistatstitle{Appendix}

\section{Hyperparameters}\label{app:hyper}
Table~\ref{tab:hyper} lists every numerical setting used in the experiments.

\begin{table}[H]\centering\small
\caption{Experimental settings for the currents-only ($d=10$) and joint currents-and-positions ($d=30$) problems.}\label{tab:hyper}
\begin{tabular}{@{}llcc@{}}
\toprule
& & currents & joint \\
\midrule
\multicolumn{2}{@{}l}{\emph{Problem}} \\
& dimension $d$                        & $10$ & $30$ \\
& tolerance $\tau$ [cm]                & \multicolumn{2}{c}{$0.5$} \\
& leg truncation margin [m]            & \multicolumn{2}{c}{$0.1$} \\
& shot / index                         & \multicolumn{2}{c}{$53349$ / $120$} \\
& current bounds [A]                   & \multicolumn{2}{c}{PX $\pm6000$, P4/P5 $\pm14400$, D1--D3 $\pm8000$,} \\
&                                      & \multicolumn{2}{c}{Dp $\pm6400$, D5 $\pm4000$, D6 $\pm3200$, D7 $\pm4800$} \\
\addlinespace
\multicolumn{2}{@{}l}{\emph{Budget}} \\
& $N_{\rm init}$                        & $50$   & $100$ \\
& matched-comparison budget             & $1000$ & $1500$ \\
& FD Jacobian solves (not counted)      & $20$   & $60$ \\
& seeds (ours / per competitor $L$)     & $50$ / $30$ & $50$ / $30$ \\
\addlinespace
\multicolumn{2}{@{}l}{\emph{Candidates}} \\
& initial pool                          & \multicolumn{2}{c}{$2\times10^{5}$} \\
& per-iteration pool                    & \multicolumn{2}{c}{$10^{5}$} \\
& L-BFGS-B refinements per iteration    & \multicolumn{2}{c}{$2$} \\
\addlinespace
\multicolumn{2}{@{}l}{\emph{Surrogates}} \\
& kernel                                & \multicolumn{2}{c}{ARD Mat\'ern-$5/2$, MLE} \\
& hyperparameter refit interval [iters] & \multicolumn{2}{c}{$10$} \\
& POD modes $K$                         & \multicolumn{2}{c}{$5$} \\
& weight of FD probes in map regression & \multicolumn{2}{c}{$1.0$ (same as an evaluation)} \\
\addlinespace
\multicolumn{2}{@{}l}{\emph{Box trust region (ours)}} \\
& initial / minimum length              & \multicolumn{2}{c}{$1.0$ / $10^{-3}$} \\
& expand factor / successes             & \multicolumn{2}{c}{$2.0$ / $2$} \\
& shrink factor / failures              & $0.5$ / $10$ & $0.5$ / $20$ \\
& axis scaling                          & \multicolumn{2}{c}{objective GP ARD lengthscales} \\
\addlinespace
\multicolumn{2}{@{}l}{\emph{Competitor}} \\
& initial length (swept, C1--C8)        & \multicolumn{2}{c}{$\{0.001, 0.005, 0.01, 0.015, 0.02, 0.025, 0.03, 0.035\}$} \\
& minimum length                        & \multicolumn{2}{c}{$10^{-4}$} \\
& initial design                        & 
$50$ (Sobol, in box, incl.\ baseline) & $100$ \\
& expand / shrink successes, failures   & \multicolumn{2}{c}{$3$ / $10$} \\
& acquisition candidates                & \multicolumn{2}{c}{$2000$ Sobol, $10$ L-BFGS-B restarts} \\
\bottomrule
\label{table:app}
\end{tabular}
\end{table}

\section{Measuring the LCFS displacement}\label{app:fd}

The baseline LCFS is the contour $\psi_0=\psi_X$, with divertor legs truncated $10$~cm beyond the X-point. To discretise this, we resample uniformly in arclength to $10^4$ points $p_k$ and compute their outward unit normals $\hat n_k$ by finite differences. 

For an equilibrium $\psi$ with poloidal magnetic flux $\psi_X'$ at the X-point, we compute $\delta C_k$ as the root $t$ nearest $0$ of $\psi(p_k+t\,\hat n_k)=\psi_X'$. This is precisely a point on the perturbed separatrix along the direction normal to our point on the LCFS. We set this to $1$~m if no such root exists within $1$~m. 

$J$, then, is the central difference of $\delta C$ with FD steps of $100$~A for currents and $2$~cm for positions ($2d$ solves). Vertices with no root at these FD probes are dropped, leaving $N\approx 10^4$ rows used throughout.

\section{Sampling details}\label{app:sampling}
Each candidate is a point $o+tu$ on a ray with origin $o\in E\cap B$, direction $u$ and length $t$.
We need to pick a origins of the ray and sample over their directions and lengths.

\emph{Origins.} We use the baseline $x_0$ (the ellipsoid centre) and the incumbent $x_{\text{best}}$, whichever lie in $E\cap B$, splitting candidates evenly between them. If neither does, we use, in order: 
\begin{itemize}
    \item the incumbent under the region snapshot that proposed it;
    \item the point $0.999$ of the way from $x_0$ to the boundary of $E$ towards $x_{\text{best}}$, if it lies in $B$;
    \item or $x_0$ but sampling $E\cap\mathcal X$ for that iteration -- that is, we drop the box trust region entirely for this iteration.
\end{itemize}

\emph{Directions.} 
To sample directions, we split evenly into two cases.

Let $M=V\Lambda V^\top$ be the eigendecomposition of the metric $M$. 
For our first half, we sample uniformly in the ellipsoid induced by the constraint pullback, a sphere in $M$. 
We take direction samples: $u=V\Lambda^{-1/2}y$, with $y=z/\|z\|$ and $z\sim\mathcal N(0,I_d)$, so that $y$ is uniform on the unit sphere. 
Since the eigenvalues of $M$ span roughly ten orders of magnitude, these directions concentrate along its small-eigenvalue eigenvectors, the coil combinations that barely move the LCFS. 

The other half are uniform Euclidean directions, which cover the remaining directions.

\emph{Lengths.} Each ray is truncated at $t_{\max}=\min(t_E,t_B)$, where $t_E$ is the positive root of $(o-x_0+tu)^\top M(o-x_0+tu)=\tau^2$ and $t_B$ the exit time from $B$. We draw $t\sim\mathrm U(0,t_{\max})$, and reduce the pool by greedy maximin.

This sampling scheme is heuristic and exact splits between the origins and direction samples will depend on the preservation constraints. It is very fast and parallelisable, however, so we can provide good coverage after down-sampling by maximin.

\section{Bilinear structure of the equilibrium response}\label{app:bilinear}

The coil (machine) contribution to the flux is linear in the coil currents,
\[
\psi_{\text{coil}}(r)
=
\sum_j I_j\,G(r;p_j),
\qquad p_j=(R_j,Z_j),
\]
where \(G\) denotes the Green's function and \(p_j\) the position of coil \(j\).

Perturbing about a reference configuration \((I^0,p^0)\) gives, for coil \(j\),
\[
\delta\psi_j
=
G_j^0\,\delta I_j
+
I_j^0\,\nabla_p G_j^0\cdot\delta p_j
+
\delta I_j\,\nabla_p G_j^0\cdot\delta p_j
+
\mathcal O(\|\delta p_j\|^2),
\]
with \(G_j^0:=G(r;p_j^0)\).
The first two terms are linear, while the third couples current and
position perturbations through the product \(\delta I_j\,\delta p_j\).

Writing \(I_j=I_j^0+\delta I_j\), the last two terms combine to \(\nabla_p G_j^0\cdot m_j\) with the flux moment \(m_j:=I_j\,\delta p_j\), so
\[
\delta\psi_j=G_j^0\,\delta I_j+\nabla_p G_j^0\cdot m_j+\mathcal O(\|\delta p_j\|^2)
\]
is linear in \(c=(\delta I,m)\).

For the joint problem, $J$ and $E$ are built in the coordinates $c$ and mapped back to $x$ (regularised for small $|I_j|$).
The refit of $\widehat J$ regresses $\delta C_i$ on $c(x_i)$,
while the remaining surrogates stay in $x$.

\end{document}